\documentclass[10pt,twocolumn,letterpaper]{article}

\usepackage{cvpr}              

\usepackage{graphicx}
\usepackage{amsmath}
\usepackage{amssymb}
\usepackage{booktabs}
\makeatletter
\@namedef{ver@everyshi.sty}{}
\makeatother
\usepackage{tikz}

\usepackage[pagebackref,breaklinks,colorlinks]{hyperref}

\usepackage[capitalize]{cleveref}
\crefname{section}{Sec.}{Secs.}
\Crefname{section}{Section}{Sections}
\Crefname{table}{Table}{Tables}
\crefname{table}{Tab.}{Tabs.}

\usepackage{booktabs}
\usepackage{multicol}
\usepackage{multirow}
\usepackage{color}
\usepackage{caption}
\usepackage{hhline}
\usepackage{pifont}
\usepackage{threeparttable}
\usepackage{makecell}
\usepackage{algorithm} 
\usepackage{listings}
\usepackage{amsfonts,amssymb}
\usepackage{graphicx}
\usepackage{lipsum}
\newcommand\blfootnote[1]{%
  \begingroup
  \renewcommand\thefootnote{}\footnote{#1}%
  \addtocounter{footnote}{-1}%
  \endgroup
}
\usepackage[accsupp]{axessibility}
\newcommand{\modelname}{PoolFormer}

\newcommand{\tabincell}[2]{\begin{tabular}{@{}#1@{}}#2\end{tabular}}

\newcommand{\myPara}[1]{\vspace{.05in}\noindent\textbf{#1.}}
\newlength\savedwidth
\newcommand\whline{\noalign{\global\savedwidth\arrayrulewidth\global\arrayrulewidth 0.8pt}\hline\noalign{\global\arrayrulewidth\savedwidth}}
\usepackage{verbatim}

\definecolor{resnet}{rgb}{0.9607843137254902, 0.8705882352941177, 0.7019607843137254}
\definecolor{regnety}{rgb}{0.9764705882352941 , 0.792156862745098 , 0.1411764705882353}
\definecolor{vit}{rgb}{0.9019607843137255 , 0.403921568627451 , 0.403921568627451}
\definecolor{deit}{rgb}{1.0 , 0.7137254901960784 , 0.7568627450980392}
\definecolor{pvt}{rgb}{0.7686274509803922 , 0.27058823529411763 , 0.4117647058823529}
\definecolor{mlpmixer}{rgb}{0.0 , 0.5803921568627451 , 0.19607843137254902}
\definecolor{resmlp}{rgb}{0.25098039215686274 , 0.8784313725490196 , 0.8156862745098039}
\definecolor{swinmixer}{rgb}{0.2196078431372549 , 0.403921568627451 , 0.8392156862745098}
\definecolor{gmlp}{rgb}{0.0 , 0.7215686274509804 , 0.5803921568627451}
\definecolor{poolformer}{rgb}{0.8666666666666667 , 0.6274509803921569 , 0.8666666666666667}
\definecolor{resnetimproved}{rgb}{1.0 , 0.54902 , 0.0}

\newcommand{\resnetdot}{\raisebox{0.5pt}{\tikz\fill[resnet] (0,0) (0ex,-1ex)--(-0.866ex,0.5ex)--(0.866ex, 0.5ex)--cycle;}}

\newcommand{\vitdot}{\raisebox{0.5pt}{\tikz\fill[vit] (0,0) (0ex,1ex)--(-0.866ex,-0.5ex)--(0.866ex, -0.5ex)--cycle;}}
\newcommand{\deitdot}{\raisebox{0.5pt}{\tikz\fill[deit] (0,0) (0ex,1ex)--(-0.866ex,-0.5ex)--(0.866ex, -0.5ex)--cycle;}}
\newcommand{\pvtdot}{\raisebox{0.5pt}{\tikz\fill[pvt] (0,0) (0ex,1ex)--(-0.866ex,-0.5ex)--(0.866ex, -0.5ex)--cycle;}}
\newcommand{\mlpmixerdot}{\raisebox{0.5pt}{\tikz\fill[mlpmixer] (0,0) (1ex,0ex)--(-0.5ex,0.866ex)--(-0.5ex, -0.866ex)--cycle;}}
\newcommand{\resmlp}{\raisebox{0.5pt}{\tikz\fill[resmlp] (0,0) (1ex,0ex)--(-0.5ex,0.866ex)--(-0.5ex, -0.866ex)--cycle;}}
\newcommand{\swinmixer}{\raisebox{0.5pt}{\tikz\fill[swinmixer] (0,0) (1ex,0ex)--(-0.5ex,0.866ex)--(-0.5ex, -0.866ex)--cycle;}}
\newcommand{\gmlp}{\raisebox{0.5pt}{\tikz\fill[gmlp] (0,0) (1ex,0ex)--(-0.5ex,0.866ex)--(-0.5ex, -0.866ex)--cycle;}}
\newcommand{\poolformer}{\raisebox{0.5pt}{\tikz\fill[poolformer] (0,0) circle (.8ex);}}

\def\cvprPaperID{3733} 
\def\confName{CVPR}
\def\confYear{2022}

\def\ie{\emph{i.e.}}

\def\etc{\emph{etc}}
\def\etal{{\em et al.~}}

\begin{document}

\title{MetaFormer Is Actually What You Need for Vision}

\author{
Weihao Yu\textsuperscript{1,2*} 
\quad Mi Luo\textsuperscript{1}
\quad Pan Zhou\textsuperscript{1}
\quad Chenyang Si\textsuperscript{1}
\quad Yichen Zhou\textsuperscript{1,2}
\\
\quad Xinchao Wang\textsuperscript{2}
\quad Jiashi Feng\textsuperscript{1}
\quad Shuicheng Yan\textsuperscript{1}
\\
\textsuperscript{1}{Sea AI Lab}
\quad \textsuperscript{2}{National University of Singapore}
\\
\small{\texttt{weihaoyu6@gmail.com} \quad \texttt{\{luomi,zhoupan,sicy,zhouyc,fengjs,yansc\}@sea.com} \quad \texttt{xinchao@nus.edu.sg}}
\\
\small{Code: \url{https://github.com/sail-sg/poolformer}}
}

\twocolumn[{
\maketitle
\vspace{-40pt}
\begin{figure}[H]
\hsize=\textwidth
\centering
\begin{subfigure}[b]{0.51\textwidth}
    \centering
    \includegraphics[width=1\textwidth]{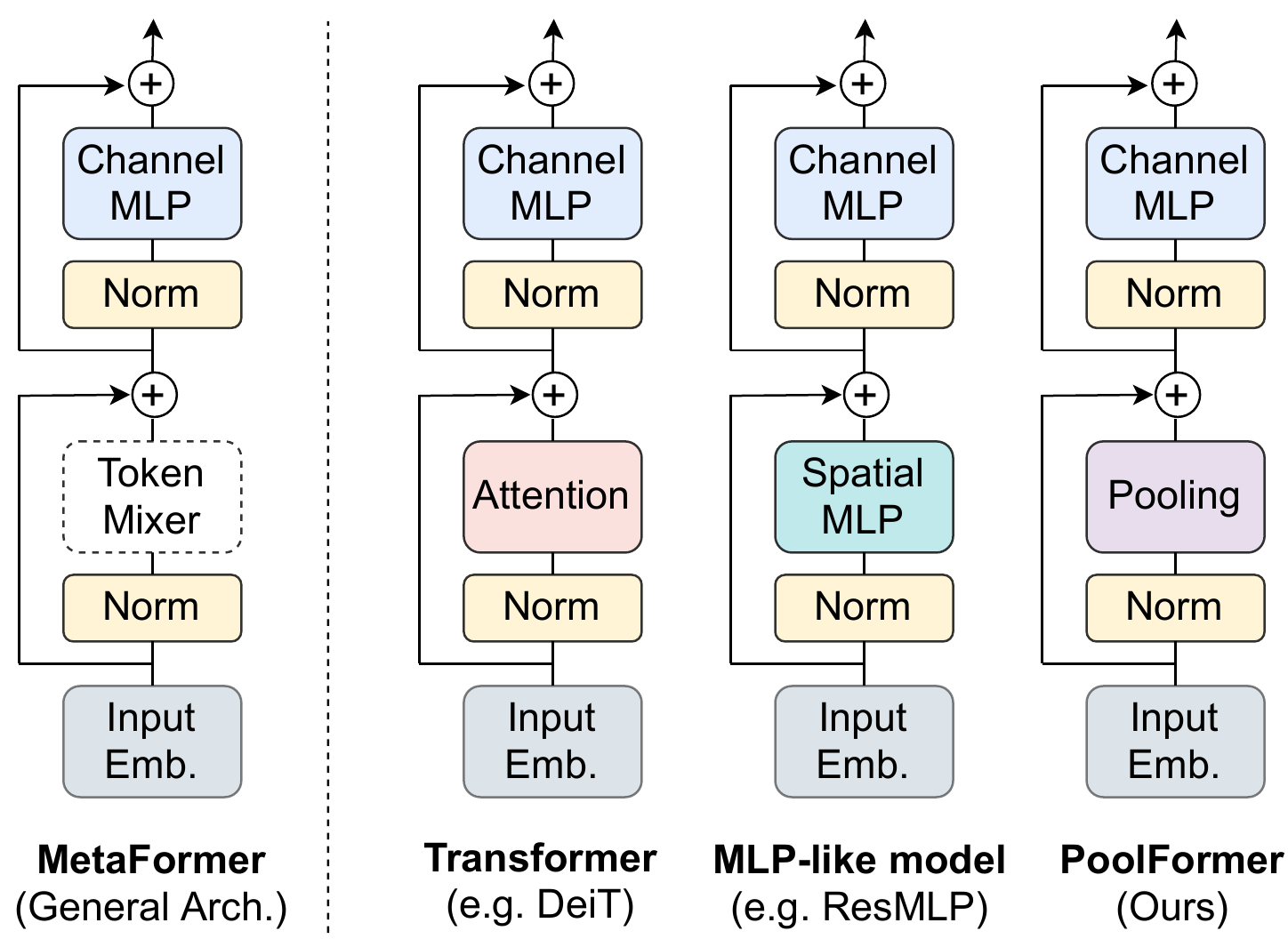}
    \caption{}
\end{subfigure}    
\hspace{0.1in}
\begin{subfigure}[b]{0.42\textwidth}
     \centering
     \includegraphics[width=1\textwidth]{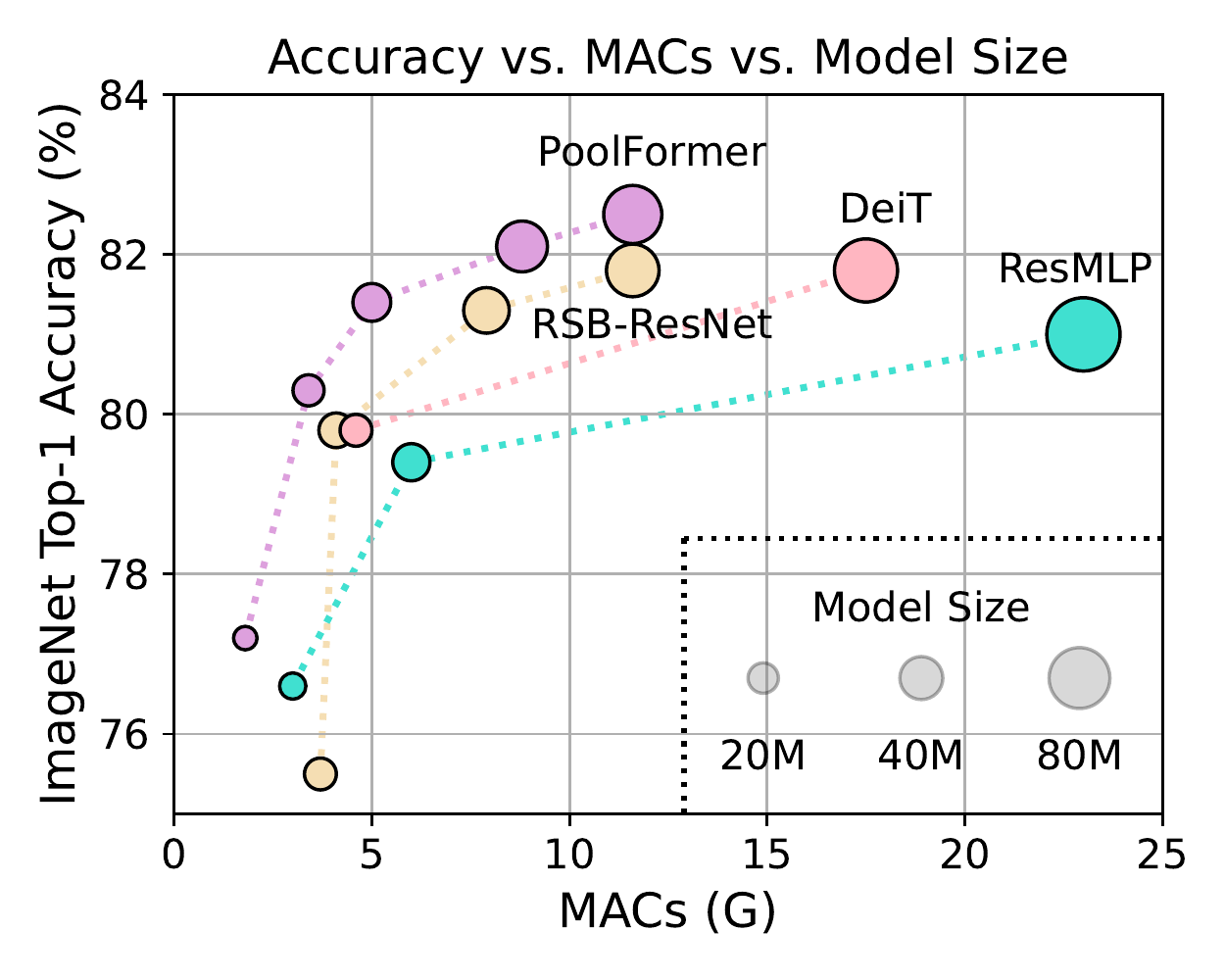}
     \caption{}
\end{subfigure}
\vspace{-2.6mm}
\caption{
\textbf{MetaFormer and performance of MetaFormer-based models on ImageNet-1K validation set.} As shown in (a),  we present \textit{MetaFormer} as a general architecture abstracted from Transformers~\cite{transformer} by not specifying the token mixer. When using attention/spatial MLP as the token mixer, MetaFormer is instantiated as Transformer/MLP-like models. We argue that the competence of Transformer/MLP-like models primarily stems from the general architecture MetaFormer instead of the equipped specific token mixers. To demonstrate this, we exploit an embarrassingly simple non-parametric operator, \textit{pooling}, to conduct extremely basic token mixing. Surprisingly, the resulted model \textit{PoolFormer} consistently outperforms the well-tuned vision Transformer \cite{vit} baseline (DeiT \cite{deit}) and MLP-like \cite{mlp-mixer} baseline (ResMLP \cite{resmlp}) as shown in (b), which well supports that MetaFormer is actually what we need to achieve competitive performance. RSB-ResNet in (b) means the results are from ``ResNet Strikes Back" \cite{resnet_improved} where ResNet \cite{resnet} are trained with improved training procedure for 300 epochs.
}
\label{fig:first_figure}
\vspace{-3mm}
\end{figure}
}]


\begin{abstract}
\vspace{-10pt}
\blfootnote{$^*$Work done during an internship at Sea AI Lab.}
Transformers have shown great potential in computer vision tasks. A common belief is their attention-based token mixer module contributes most to their competence. However, recent works show the attention-based module in Transformers can be replaced by spatial MLPs and the resulted models still perform quite well. Based on this observation, we hypothesize that the general architecture of the Transformers, instead of the specific token mixer module, is more essential to the model's performance. To verify this, we deliberately replace the attention module in Transformers with an embarrassingly simple \emph{spatial pooling} operator to conduct only basic token mixing. Surprisingly, we observe that the derived model, termed as PoolFormer, achieves competitive performance on multiple computer vision tasks. For example, on ImageNet-1K, PoolFormer achieves 82.1\% top-1 accuracy, surpassing well-tuned Vision Transformer/MLP-like baselines DeiT-B/ResMLP-B24 by 0.3\%/1.1\% accuracy with 35\%/52\% fewer parameters and 50\%/62\% fewer MACs. The effectiveness of PoolFormer verifies our hypothesis and urges us to initiate the concept of ``MetaFormer", a general architecture abstracted from Transformers without specifying the token mixer. Based on the extensive experiments, we argue that MetaFormer is the key player in achieving superior results for recent Transformer and MLP-like models on vision tasks. This work calls for more future research dedicated to improving MetaFormer instead of focusing on the token mixer modules. Additionally, our proposed PoolFormer could serve as a starting baseline for future MetaFormer architecture design.
\end{abstract}

\vspace{-30pt}
\section{Introduction}
Transformers have gained much interest and success in the computer vision field~\cite{double_attention, stand_alone_attention, vaswani2021scaling, detr}.  
Since the seminal work of Vision Transformer (ViT) \cite{vit} that adapts pure Transformers to image classification tasks, many follow-up models are developed to make further improvements and achieve promising performance in various computer vision tasks \cite{deit, t2t, swin}.

The Transformer encoder, as shown in Figure \ref{fig:first_figure}(a), consists of two  components. One is the attention module for mixing information among tokens and we term it as \textit{token mixer}. The other component contains the remaining modules, such as channel MLPs and residual connections. By regarding the attention module as a specific token mixer, we further abstract the overall Transformer into a general architecture \textit{MetaFormer} where the token mixer is not specified, as shown in Figure \ref{fig:first_figure}(a).

The success of Transformers has been long attributed to the attention-based token mixer \cite{transformer}. Based on this common belief, many variants of the attention modules  \cite{convit, pvt, refiner, tnt} have been developed to improve the Vision Transformer. However, a very recent work~\cite{mlp-mixer}  replaces the attention module completely with spatial MLPs as token mixers, and finds the derived  MLP-like model can readily attain competitive performance on image classification benchmarks. The follow-up works \cite{resmlp, gmlp, vip} further improve MLP-like models by data-efficient training and specific MLP module design, gradually narrowing the performance gap to ViT and challenging the dominance of attention as token mixers.

Some recent approaches~\cite{fnet, infinite_former, gfnet, continus_attention} explore other types of token mixers within the MetaFormer architecture, and have demonstrated encouraging performance. For example, \cite{fnet} replaces attention with Fourier Transform and still achieves around 97\% of the accuracy of vanilla Transformers. Taking all these results together, it seems as long as a model adopts MetaFormer as the general architecture, promising results could be attained. We thus hypothesize \textit{compared with specific token mixers, MetaFormer is more essential for the model to achieve competitive performance}.

To verify this hypothesis, we apply an extremely simple non-parametric operator, \emph{pooling}, as the token mixer to conduct only basic token mixing. Astonishingly, this derived model, termed \emph{PoolFormer}, achieves competitive performance, and even consistently outperforms well-tuned Transformer and MLP-like models, including DeiT \cite{deit} and ResMLP \cite{resmlp}, as shown in Figure \ref{fig:first_figure}(b). More specifically, PoolFormer-M36 achieves 82.1\% top-1 accuracy on ImageNet-1K classification benchmark, surpassing well-tuned vision Transformer/MLP-like baselines DeiT-B/ResMLP-B24 by 0.3\%/1.1\% accuracy with 35\%/52\% fewer parameters and 50\%/62\% fewer MACs. These results demonstrate that MetaFormer, even with a naive token mixer, can still deliver promising performance. We thus argue that MetaFormer is our \emph{de facto} need for vision models which is more essential to achieve competitive performance rather than specific token mixers. Note that it does not mean the token mixer is insignificant. MetaFormer still has this abstracted component. It means token mixer is not limited to a specific type, e.g. attention.

The contributions of our paper are two-fold. Firstly, we abstract Transformers into a general architecture MetaFormer, and empirically demonstrate that the success of Transformer/MLP-like models is largely attributed to the  MetaFormer architecture. Specifically, by only employing a simple non-parametric operator, pooling, as an extremely weak token mixer for MetaFormer, we build a simple model named PoolFormer and find it can still achieve highly competitive performance. We hope our findings inspire more future research dedicated to improving MetaFormer instead of focusing on the token mixer modules. Secondly, we evaluate the proposed PoolFormer on multiple vision tasks including image classification \cite{imagenet}, object detection \cite{coco}, instance segmentation \cite{coco}, and semantic segmentation \cite{ade20k}, and find it achieves competitive performance compared with the SOTA models using sophistic design of token mixers. The PoolFormer can readily serve as a good starting baseline for future MetaFormer architecture design.

\section{Related work}
Transformers are first proposed by \cite{transformer} for translation tasks and then rapidly become popular in various NLP tasks. 
In language pre-training tasks, Transformers are trained on large-scale unlabeled text corpus and achieve amazing performance \cite{bert, gpt3}. 
Inspired by the success of Transformers in NLP, many researchers apply attention mechanism and Transformers to vision tasks \cite{double_attention, stand_alone_attention, vaswani2021scaling, detr}. Notably, Chen \etal introduce iGPT \cite{igpt} where the Transformer is trained to auto-regressively predict pixels on images for self-supervised learning. Dosovitskiy \etal propose Vision Transformer (ViT) with hard patch embedding as input\cite{vit}. They show that on supervised image classification tasks, a ViT pre-trained on a large propriety dataset (JFT dataset with 300 million images) can achieve excellent performance. DeiT \cite{deit} and T2T-ViT \cite{t2t} further demonstrate that the ViT pre-trained on only ImageNet-1K ($\sim$ 1.3 million images) from scratch can achieve promising performance. A lot of works have been focusing on improving the token mixing approach of Transformers by shifted windows~\cite{swin}, relative position encoding~\cite{wu2021rethinking}, refining attention map~\cite{refiner}, or incorporating convolution \cite{guo2021cmt, wu2021cvt, d2021convit}, \etc. In addition to attention-like token mixers, \cite{mlp-mixer, resmlp} surprisingly find that merely adopting MLPs as token mixers can still achieve competitive performance. This discovery challenges the dominance of attention-based token mixers and triggers a heated discussion in the research community about which token mixer is better \cite{vip, chen2021cyclemlp}. However, the target of this work is neither to be engaged in this debate nor to design new complicated token mixers to achieve new state of the art. Instead, we examine a fundamental question: What is truly responsible for the success of the Transformers and their variants? Our answer is the general architecture \ie, MetaFormer. We simply utilize pooling as basic token mixers to probe the power of MetaFormer.

Contemporarily, some works contribute to answering the same question. Dong \etal prove that without residual connections or MLPs, the output converges doubly exponentially to a rank one matrix \cite{dong2021attention}. Raghu \etal \cite{raghu2021vision} compare the feature difference between ViT and CNNs, finding that self-attention allows early gathering of global information while residual connections greatly propagate features from lower layers to higher ones. Park \etal \cite{park2022how} shows that multi-head self-attentions improve accuracy and generalization by flattening the loss landscapes. Unfortunately, they do not abstract Transformers into a general architecture and study them from the aspect of general framework.
\section{Method}
\begin{figure*}[t]
  \centering
   \includegraphics[width=0.85\linewidth]{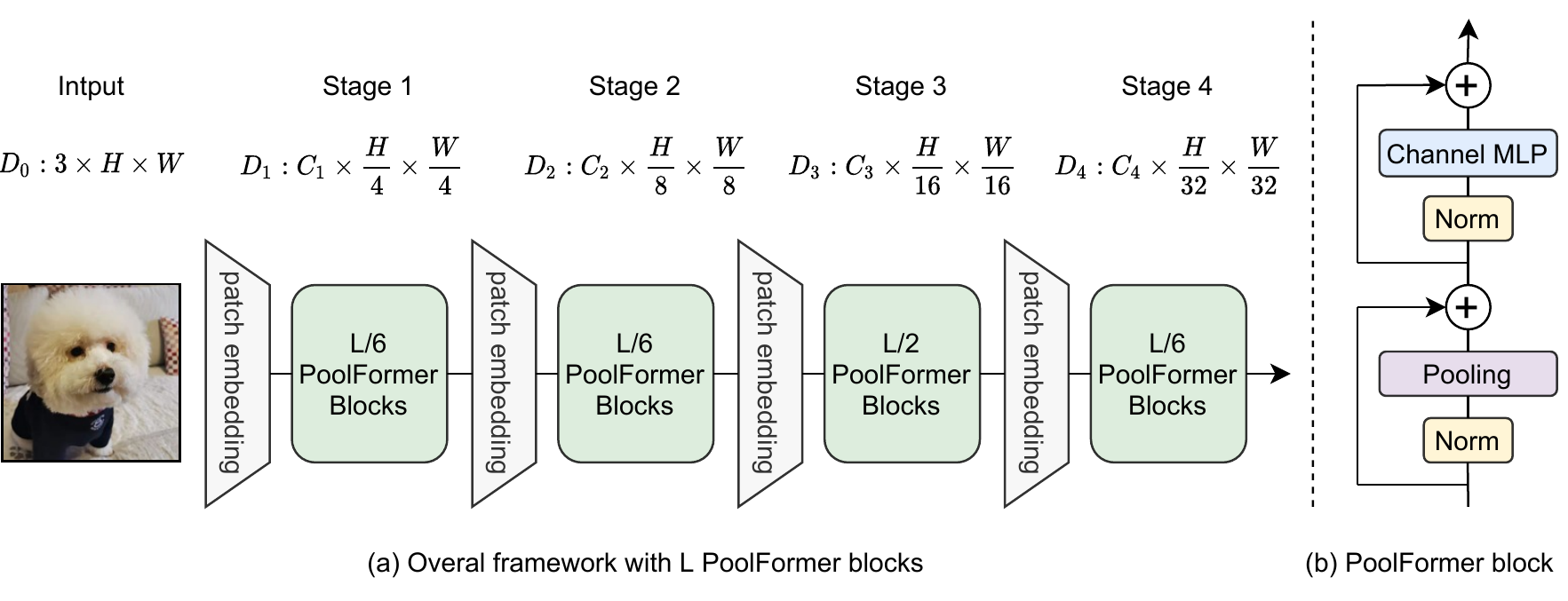}
   \vspace{-4mm}
   \caption{\textbf{(a) The overall framework of \modelname{}.} Similar to \cite{resnet, pvt, swin}, \modelname{} adopts hierarchical architecture with 4 stages. For a model with L \modelname{} blocks, stage [1, 2, 3, 4] have [L/6, L/6, L/2, L/6] blocks, respectively. The feature dimension $D_i$ of stage $i$ is shown in the figure. \textbf{(b) The architecture of \modelname{} block.} Compared with Transformer block, it replaces attention with extremely simple non-parametric operator, pooling, to conduct only basic token mixing.}
   \label{fig:overall_architecture}
\end{figure*}

\begin{algorithm}[t]
\caption{Pooling for PoolFormer, PyTorch-like Code}
\label{alg:code}
\definecolor{codeblue}{rgb}{0.25,0.5,0.5}
\definecolor{codekw}{rgb}{0.85, 0.18, 0.50}
\lstset{
  backgroundcolor=\color{white},
  basicstyle=\fontsize{7.5pt}{7.5pt}\ttfamily\selectfont,
  columns=fullflexible,
  breaklines=true,
  captionpos=b,
  commentstyle=\fontsize{7.5pt}{7.5pt}\color{codeblue},
  keywordstyle=\fontsize{7.5pt}{7.5pt}\color{codekw},
}
\begin{lstlisting}[language=python]
import torch.nn as nn

class Pooling(nn.Module):
    def __init__(self, pool_size=3):
        super().__init__()
        self.pool = nn.AvgPool2d(
            pool_size, stride=1, 
            padding=pool_size//2, 
            count_include_pad=False,
        )
    def forward(self, x):
        """
        [B, C, H, W] = x.shape
        Subtraction of the input itself is added 
        since the block already has a 
        residual connection.
        """
        return self.pool(x) - x
\end{lstlisting}
\end{algorithm}

\subsection{MetaFormer}
We present the core concept ``MetaFormer" for this work at first. As shown in Figure \ref{fig:first_figure}, abstracted from Transformers \cite{transformer}, 
MetaFormer is a general architecture where the token mixer is not specified while the other components are kept the same as Transformers. The input $I$ is first processed by input embedding, such as  patch embedding for ViTs \cite{vit},
\begin{equation}
    X = \mathrm{InputEmb}(I),
\end{equation}
where  $X \in \mathbb{R}^{N \times C}$ denotes the embedding tokens with sequence length $N$ and embedding dimension $C$.

Then, embedding tokens are fed to repeated MetaFormer blocks, each of which includes two residual sub-blocks. Specifically, the first sub-block mainly contains a token mixer to communicate information among tokens and this sub-block can be expressed as
\begin{equation}
    Y = \mathrm{TokenMixer}(\mathrm{Norm}(X)) + X,
\end{equation}
where $\mathrm{Norm}(\cdot)$ denotes the normalization such as Layer Normalization \cite{layer_norm} or Batch Normalization \cite{batch_norm}; $\mathrm{TokenMixer}(\cdot)$ means a module mainly working for mixing token information. It is implemented by various attention mechanism in recent vision Transformer models  \cite{vit,refiner,t2t} or spatial MLP in MLP-like models \cite{mlp-mixer, resmlp}. Note that the main function of the token mixer is to propagate token information although some token mixers can also mix channels, like attention.

The second sub-block primarily consists of a two-layered MLP with non-linear activation, 
\begin{equation}
    Z = \sigma(\mathrm{Norm}(Y)W_1)W_2 + Y,
\end{equation}
where $W_1 \in \mathbb{R}^{C \times rC}$ and $W_2 \in \mathbb{R}^{rC \times C}$ are learnable parameters with MLP expansion ratio $r$; $\sigma(\cdot)$ is a non-linear activation function, such as GELU \cite{gelu} or ReLU \cite{relu}. 

\myPara{Instantiations of MetaFormer} MetaFormer describes a general architecture 
with which different models can be obtained immediately by specifying the concrete design of the token mixers. 
As shown in Figure \ref{fig:first_figure}(a), if the token mixer is specified as attention or spatial MLP, MetaFormer then becomes a Transformer or MLP-like model respectively. 

\subsection{PoolFormer}
From the introduction of Transformers \cite{transformer}, lots of works attach much importance to the attention and focus on designing various attention-based token mixer components. In contrast, these works pay little attention to the general architecture, \ie, the MetaFormer.

In this work, we argue that this MetaFormer general architecture contributes mostly to the success of the recent Transformer and MLP-like models. 
To demonstrate it, we deliberately employ an embarrassingly simple operator, pooling, as the token mixer. This operator has no learnable parameters and it just makes each token averagely aggregate its nearby token features.

Since this work is targeted at vision tasks,  we assume the input is in channel-first data format, \ie,  $T \in \mathbb{R}^{C \times H \times W}$. The pooling operator can be expressed as
\begin{equation}
\label{eq:pool}
    T'_{:, i, j} =  \frac{1}{K \times K} \sum_{p,q=1}^{K}T_{:, i+p-\frac{K+1}{2}, i+q-\frac{K+1}{2}} - T_{:, i, j},
\end{equation}
where $K$ is the pooling size. Since the MetaFormer block already has a residual connection, subtraction of the input itself is added in Equation (\ref{eq:pool}). The PyTorch-like code of the pooling is shown in Algorithm \ref{alg:code}.

As well known, self-attention and spatial MLP have computational complexity quadratic to the number of tokens to mix. Even worse, spatial MLPs bring much more parameters when handling longer sequences. As a result, self-attention and spatial MLPs usually can only process hundreds of tokens. In contrast, the pooling needs a computational complexity linear to the sequence length without any learnable parameters.  Thus, we take advantage of pooling by adopting a hierarchical structure similar to traditional CNNs \cite{alexnet, vgg, resnet} and recent hierarchical Transformer variants \cite{swin, pvt}. Figure \ref{fig:overall_architecture} shows the overall framework of PoolFormer. Specifically, PoolFormer has 4 stages with $\frac{H}{4} \times \frac{W}{4}$, $\frac{H}{8} \times \frac{W}{8}$, $\frac{H}{16} \times \frac{W}{16}$, and $\frac{H}{32} \times \frac{W}{32}$ tokens respectively, where $H$ and $W$ represent the width and height of the input image. There are two groups of embedding size: 1) small-sized models with embedding dimensions of 64, 128, 320, and 512 responding to the four stages; 2) medium-sized models with embedding dimensions 96, 192, 384, and 768. Assuming there are $L$ PoolFormer blocks in total, stages 1, 2, 3, and 4 will contain $L/6$, $L/6$, $L/2$, and $L/6$ PoolFormer blocks respectively. The MLP expansion ratio is set as 4. According to the above simple model scaling rule, we obtain 5 different model sizes of PoolFormer and their hyper-parameters are shown in Table \ref{tab:model}.

\begin{table}[t]
\footnotesize
\centering
\setlength{\tabcolsep}{2pt}
\newcommand{\blockc}[4]{
$\begin{bmatrix}
	\begin{array}{l}
	R_1=#1 \\
	N_1=#2 \\
	E_1=#3 \\
	\end{array}
\end{bmatrix} \times #4$
}

\newcommand{\sblock}[3]{
$\begin{matrix}
E_{#1}=#2 \\
L_{#1}=#3 \\
\end{matrix}$
}

\newcommand{\poollayer}{
Pooling Size & \multicolumn{5}{c}{$3 \times 3$, stride 1}\\
\cline{4-9}
}

\newcommand{\stitle}[6]{
\multirow{5}{*}{#1} & \multirow{5}{*}{\scalebox{1}{$\frac{H}{#2}\times \frac{W}{#2}$}} & \multirow{2}{*}{\tabincell{c}{Patch \\ Embedding}} & Patch Size & \multicolumn{5}{c}{$#3 \times #3$, stride $#4$} \\
\cline{4-9}
    &    &    & Embed. Dim. & \multicolumn{3}{c|}{$#5$} & \multicolumn{2}{c}{$#6$} \\
\cline{3-9}
& & \multirow{3}{*}{\tabincell{c}{\modelname{}\\Block}} 
}

\begingroup
\renewcommand{\arraystretch}{1.1}
\begin{tabular}{c|c|c|c|c|c|c|c|c}
\toprule
  \multirow{2}{*}{Stage} & \multirow{2}{*}{\#Tokens} & \multicolumn{2}{c|}{\multirow{2}{*}{Layer Specification}} & \multicolumn{5}{c}{\modelname{}} \\
\cline{5-9}
 & & \multicolumn{2}{c|}{} & S12 & S24 & S36 & M36 & M48 \\
\whline
\stitle{1}{4}{7}{4}{64}{96}    & \poollayer
 & & & MLP Ratio & \multicolumn{5}{c}{4} \\
\cline{4-9}
 & & & \# Block & 2 & 4 & 6 & 6 & 8 \\
\hline
\stitle{2}{8}{3}{2}{128}{192}  & \poollayer
 & & & MLP Ratio & \multicolumn{5}{c}{4} \\
 \cline{4-9}
 & & & \# Block & 2 & 4 & 6 & 6 & 8 \\
\hline
\stitle{3}{16}{3}{2}{320}{384} & \poollayer
 & & & MLP Ratio & \multicolumn{5}{c}{4} \\
\cline{4-9}
 & & & \# Block & 6 & 12 & 18 & 18 & 24 \\
\hline
\stitle{4}{32}{3}{2}{512}{768} & \poollayer
 & & & MLP Ratio & \multicolumn{5}{c}{4} \\
 \cline{4-9}
 & & & \# Block & 2 & 4 & 6 & 6 & 8 \\
\hline
\multicolumn{4}{c|}{Parameters~(M)}& 11.9 & 21.4              &  30.8            &  56.1              &  73.4    \\
\hline
\multicolumn{4}{c|}{MACs~(G)}     & 1.8  &  3.4              &  5.0             &  8.8              &  11.6    \\
\bottomrule
\end{tabular}
\endgroup

\vspace{-3mm}
\caption{\textbf{ Configurations of different PoolFormer models.} There are two groups of embedding dimensions, \ie, small size with [64, 128, 320, 512] dimensions and medium size with [96, 196, 384, 768]. Notation ``S24" means the model is in small size of embedding dimensions with 24 PoolFormer blocks in total. The numbers of MACs are counted by \texttt{fvcore}\cite{fvcore} library.
}
\label{tab:model}
\vspace{-4mm}
\end{table}

\vspace{-5pt}
\section{Experiments}

\begin{table*}[t]
    \centering
    \small
    
    \begin{tabular}{c|clccccccc}
        \toprule
      General Arch.  &  Token Mixer    & Outcome Model    & Image Size & Params (M)  & MACs (G) & Top-1 (\%) \\
        \whline
    {\multirow{5}{*}{\makecell[c]{Convolutional \\ Neural Netowrks}}}  & \multirow{5}{*}{---}   & \resnetdot{} RSB-ResNet-18 \cite{resnet, resnet_improved} & 224 & 12 & 1.8 & 70.6 \\
         &   & \resnetdot{} RSB-ResNet-34 \cite{resnet, resnet_improved} & 224 & 22 & 3.7 & 75.5 \\
         &   & \resnetdot{} RSB-ResNet-50 \cite{resnet, resnet_improved} & 224 & 26 & 4.1 & 79.8 \\
         &   & \resnetdot{} RSB-ResNet-101 \cite{resnet, resnet_improved} & 224 &  45 & 7.9 & 81.3 \\
         &   & \resnetdot{} RSB-ResNet-152 \cite{resnet, resnet_improved} & 224 & 60 & 11.6 & 81.8 \\
        \hline
         
   \multirow{22}{*}{MetaFormer}  & \multirow{8}{*}{Attention} & \vitdot{} ViT-B/16$^*$ \cite{vit} & 224 &  86 & 17.6 & 79.7 \\
             &          & \vitdot{} ViT-L/16$^*$ \cite{vit} & 224 & 307 & 63.6 & 76.1 \\
             &          & \deitdot{} DeiT-S \cite{deit} & 224 & 22 & 4.6 & 79.8 \\
             &          & \deitdot{} DeiT-B \cite{deit} & 224 &  86 & 17.5 & 81.8 \\
             &          & \pvtdot{} PVT-Tiny \cite{pvt} & 224 & 13 & 1.9 & 75.1 \\
             &          & \pvtdot{} PVT-Small \cite{pvt} & 224 & 25 & 3.8 & 79.8 \\
             &          & \pvtdot{} PVT-Medium \cite{pvt} & 224 & 44 &  6.7 & 81.2 \\
             &          & \pvtdot{} PVT-Large \cite{pvt} & 224 & 61 &  9.8 & 81.7 \\
    \cline{2-7}
             & \multirow{9}{*}{Spatial MLP} & \mlpmixerdot{} MLP-Mixer-B/16 \cite{mlp-mixer} & 224 & 59 & 12.7 & 76.4 \\
             &             & \resmlp{} ResMLP-S12 \cite{resmlp} & 224 & 15 & 3.0 & 76.6 \\
             &             & \resmlp{} ResMLP-S24 \cite{resmlp} & 224 & 30 & 6.0 & 79.4 \\
             &             & \resmlp{} ResMLP-B24 \cite{resmlp} & 224 & 116 & 23.0 & 81.0 \\
             &             & \swinmixer{} Swin-Mixer-T/D24 \cite{swin} & 256 & 20 & 4.0 & 79.4 \\
             &             & \swinmixer{} Swin-Mixer-T/D6 \cite{swin} & 256 & 23 & 4.0 & 79.7 \\
             &             & \swinmixer{} Swin-Mixer-B/D24 \cite{swin} & 224 & 61 & 10.4 & 81.3 \\
             &             & \gmlp{} gMLP-S \cite{gmlp} & 224 & 20 & 4.5 & 79.6 \\
             &             & \gmlp{} gMLP-B \cite{gmlp} & 224 & 73 & 15.8 & 81.6 \\
    \cline{2-7}
             & \multirow{5}{*}{Pooling}  & \poolformer{} PoolFormer-S12 & 224 & 12 & 1.8 & 77.2 \\ 
             &         & \poolformer{} PoolFormer-S24 & 224 & 21 & 3.4 & 80.3 \\
             &         & \poolformer{} PoolFormer-S36 & 224 & 31 & 5.0 & 81.4 \\ 
             &         & \poolformer{} PoolFormer-M36 & 224 & 56 & 8.8 & 82.1 \\
             &         & \poolformer{} PoolFormer-M48 & 224 & 73 & 11.6 & 82.5 \\
    \bottomrule
             
    \end{tabular}
    \vspace{-3mm}
    \caption{
    \textbf{Performance of different types of models on ImageNet-1K  classification.} 
    All these models are only trained on the ImageNet-1K training set and the accuracy on the validation set is reported. RSB-ResNet means the results are from ``ResNet Strikes Back" \cite{resnet_improved} where ResNet \cite{resnet} is trained with improved training procedure for 300 epochs. $^*$ denotes results of ViT trained with extra regularization from \cite{mlp-mixer}. The numbers of MACs of PoolFormer are counted by \texttt{fvcore} \cite{fvcore} library.
    \label{tab:imagenet}}
    \vspace{-2mm}
\end{table*}

\begin{figure*}[t]
  \centering
   \includegraphics[width=0.95\linewidth]{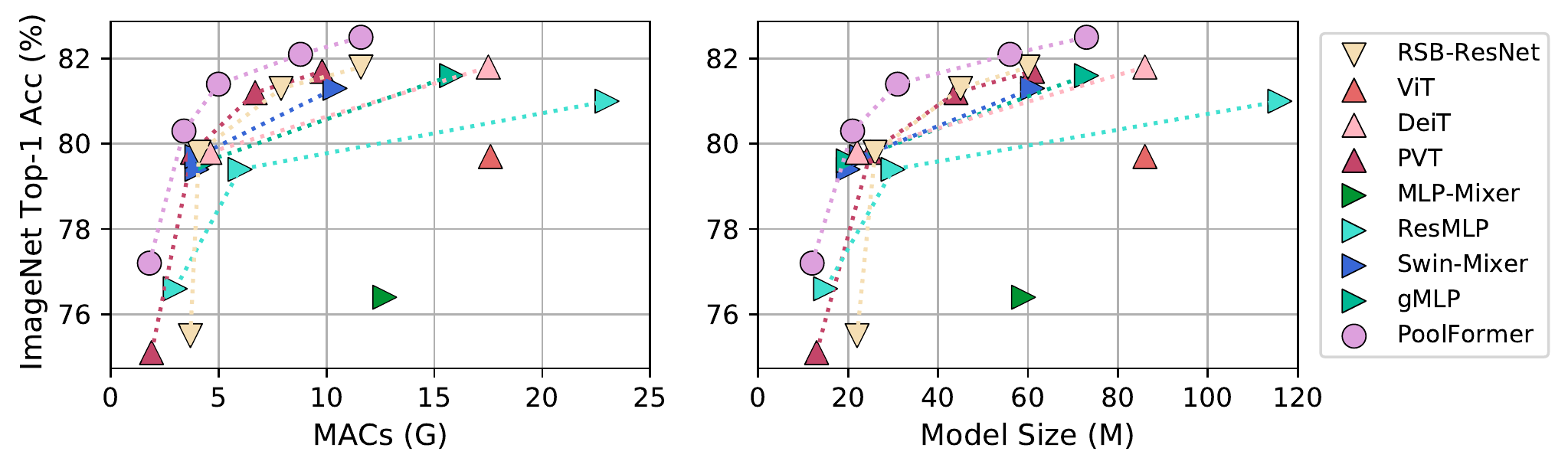}
   \vspace{-4mm}
   \caption{\textbf{ImageNet-1K validation accuracy \vs MACs/Model Size.} RSB-ResNet means the results are from ``ResNet Strikes Back" \cite{resnet_improved} where ResNet \cite{resnet} is trained with improved training procedure for 300 epochs.}
   \label{fig:overall_comparision}
\vspace{-1mm}
\end{figure*}

\subsection{Image classification}
\myPara{Setup} 
ImageNet-1K \cite{imagenet} is one of the most widely used datasets in computer vision. It contains about 1.3M training images and 50K validation images, covering common 1K classes. Our training scheme mainly follows \cite{deit} and \cite{cait}. Specifically, MixUp \cite{mixup}, CutMix \cite{cutmix}, CutOut \cite{cutout} and RandAugment \cite{randaugment} are used for data augmentation. The models are trained for 300 epochs using AdamW optimizer \cite{adam, adamw} with weight decay $0.05$ and peak learning rate $\mathrm{lr} = 1e^{-3} \cdot \mathrm{batch\ size} / 1024$ (batch size 4096 and learning rate $4e^{-3}$ are used in this paper). The number of warmup epochs is 5 and cosine schedule is used to decay the learning rate. Label Smoothing \cite{label_smoothing} is set as 0.1. Dropout is disabled but stochastic depth \cite{stochastic_depth} and LayerScale \cite{cait} are used to help train deep models. 
We modified Layer Normalization \cite{layer_norm} to compute the mean and variance along token and channel dimensions compared to only channel dimension in vanilla Layer Normalization. Modified Layer Normalization (MLN) can be implemented for channel-first data format with GroupNorm API in PyTorch by specifying the group number as 1. MLN is preferred by PoolFormer as shown in Section \ref{sec:ablation}. See the appendix for more details on hyper-parameters. Our implementation is based on the \texttt{Timm} codebase \cite{timm} and the experiments are run on TPUs.

\myPara{Results} 
Table \ref{tab:imagenet} shows the performance of PoolFormers on ImageNet classification. Qualitative results are shown in the appendix. Surprisingly, despite the simple pooling token mixer, PoolFormers can still achieve highly competitive performance compared with CNNs and other MetaFormer-like models. For example, PoolFormer-S24 reaches the top-1 accuracy of more than 80 while only requiring 21M parameters and 3.4G MACs. Comparatively, the well-established ViT baseline DeiT-S \cite{deit}, attains slightly worse accuracy of 79.8 and requires 35\% more MACs (4.6G). To obtain similar accuracy, MLP-like model ResMLP-S24 \cite{resmlp} needs 43\% more parameters (30M) as well as 76\% more computation (6.0G) while only 79.4 accuracy is attained. Even compared with more improved ViT and MLP-like variants \cite{pvt, gmlp}, PoolFormer still shows better performance. Specifically, the pyramid Transformer PVT-Medium obtains 81.2 top-1 accuracy with 44M parameters and 6.7G MACs while PoolFormer-S36 reaches 81.4 with 30\% fewer parameters (31M) and 25\% fewer MACs (5.0G) than those of PVT-Medium.

Besides, compared with RSB-ResNet (``ResNet Strikes Back") \cite{resnet_improved} where ResNet \cite{resnet} is trained with improved training procedure for the same 300 epochs, PoolFormer still performs better. With $\sim$ 22M parameters/3.7G MACs, RSB-ResNet-34 \cite{resnet_improved} gets 75.5 accuracy while PoolFormer-S24 can obtain 80.3. Since the local spatial modeling ability of the pooling layer is much worse than the neural convolution layer, the competitive performance of PoolFormer can only be attributed to its general architecture MetaFormer.

With the pooling operator,  each token evenly aggregates the features from its nearby tokens. Thus it is an extremely basic token mixing operation. However, the experiment results show that even with this embarrassingly simple token mixer, MetaFormer still obtains highly competitive performance. 
Figure \ref{fig:overall_comparision} clearly shows that PoolFormer surpasses other models with fewer MACs and parameters. 
This finding conveys that the general architecture MetaFormer is actually what we need when designing vision models. By adopting MetaFormer, it is guaranteed that the derived models would have the potential to achieve reasonable performance. 

\begin{table*}[t]
\small
\centering
\setlength{\tabcolsep}{2pt}
\begin{tabular}{l|c|ccc|ccc|c|ccc|ccc}
\toprule
\multirow{2}{*}{Backbone} & \multicolumn{7}{c|}{RetinaNet 1$\times$} & \multicolumn{7}{c}{Mask R-CNN 1$\times$} \\
\cline{2-15}
& Params (M) & AP & AP$_{50}$ &AP$_{75}$ & AP$_S$ & AP$_M$ & AP$_L$ & Params (M) & AP$^{\rm b}$ &AP$_{50}^{\rm b}$ &AP$_{75}^{\rm b}$  &AP$^{\rm m}$ &AP$_{50}^{\rm m}$ &AP$_{75}^{\rm m}$ \\
\whline
\resnetdot{} ResNet-18~\cite{resnet}  & 21.3 & 31.8 & 49.6 & 33.6 & 16.3 & 34.3 & 43.2 & 31.2 &  34.0 & 54.0 & 36.7 & 31.2 & 51.0 & 32.7 \\
\poolformer{} PoolFormer-S12          & 21.7 & 36.2 & 56.2 & 38.2 & 20.8 & 39.1 & 48.0 & 31.6 &  37.3 & 59.0 & 40.1 & 34.6 & 55.8 & 36.9 \\
\hline
\resnetdot{} ResNet-50~\cite{resnet}  & 37.7 & 36.3 & 55.3 & 38.6 & 19.3 & 40.0 & 48.8 & 44.2 &  38.0 & 58.6 & 41.4 & 34.4 & 55.1 & 36.7 \\
\poolformer{} PoolFormer-S24          & 31.1 & 38.9 & 59.7 & 41.3 & 23.3 & 42.1 & 51.8 & 41.0 &  40.1 & 62.2 & 43.4 & 37.0 & 59.1 & 39.6 \\
\hline
\resnetdot{} ResNet-101~\cite{resnet}  & 56.7 & 38.5 & 57.8 & 41.2 & 21.4 & 42.6 & 51.1 & 63.2 &  40.4 & 61.1 & 44.2 & 36.4 & 57.7 & 38.8 \\
\poolformer{} PoolFormer-S36           & 40.6 & 39.5 & 60.5 & 41.8 & 22.5 & 42.9 & 52.4 & 50.5 &  41.0 & 63.1 & 44.8 & 37.7 & 60.1 & 40.0 \\ 
\bottomrule
\end{tabular}
\vspace{-3mm}
\caption{\textbf{Performance of object detection using RetinaNet, and object detection and instance segmentation using Mask R-CNN on COCO \texttt{val2017}~\cite{coco}.} $1 \times$ training schedule (\ie 12 epochs) is used for training detection models. $AP^b$ and $AP^m$ represent bounding box AP and mask AP, respectively.}
\label{tab:coco_det} 
\vspace{-3mm}
\end{table*}

\begin{table}[t]
\small
\centering
\setlength{\tabcolsep}{3.5pt}
\begin{tabular}{l|c|c}
\toprule
\multirow{2}{*}{Backbone} & \multicolumn{2}{c}{Semantic FPN}\\
\cline{2-3}
& Params (M) & mIoU (\%) \\
    \whline
	\resnetdot{} ResNet-18~\cite{resnet}                & 15.5 &  32.9 \\
	\pvtdot{} PVT-Tiny~\cite{pvt}           & 17.0 &  35.7 \\
	\poolformer{} PoolFormer-S12                    & 15.7 &  37.2 \\
	\hline
    \resnetdot{} ResNet-50~\cite{resnet}                & 28.5 &  36.7 \\
    \pvtdot{} PVT-Small~\cite{pvt}          & 28.2 &  39.8 \\
	\poolformer{} PoolFormer-S24                    & 23.2 &  40.3 \\
    \hline
    \resnetdot{} ResNet-101~\cite{resnet}               & 47.5 &  38.8\\
    \resnetdot{} ResNeXt-101-32x4d~\cite{xie2017aggregated} & 47.1 &  39.7 \\
    \pvtdot{} PVT-Medium~\cite{pvt}         & 48.0 & 41.6 \\
    \poolformer{} PoolFormer-S36                    & 34.6 & 42.0 \\
    \hline
    \pvtdot{} PVT-Large~\cite{pvt}          & 65.1 &  42.1 \\
    \poolformer{} PoolFormer-M36                    & 59.8 & 42.4 \\
    \hline
    \resnetdot{} ResNeXt-101-64x4d~\cite{xie2017aggregated} & 86.4 &  40.2 \\
    \poolformer{} PoolFormer-M48                   & 77.1 &  42.7 \\
\bottomrule
\end{tabular}
\vspace{-3mm}
\caption{\textbf{Performance of Semantic segmentation on ADE20K~\cite{ade20k} validation set.} All models are equipped with Semantic FPN~\cite{fpn}.}
\label{tab:ade20k}
\normalsize
\vspace{-5mm}
\end{table}

\subsection{Object detection and instance segmentation}
\vspace{-1pt}
\myPara{Setup} We evaluate PoolFormer on the challenging COCO benchmark \cite{coco} that includes 118K training images (\texttt{train2017}) and 5K validation images (\texttt{val2017}). The models are trained on training set and the performance on validation set is reported. PoolFormer is employed as the backbone for two standard detectors, \ie, RetinaNet~\cite{retinanet} and Mask R-CNN~\cite{mask_rcnn}. ImageNet pre-trained weights are utilized to initialize the backbones and Xavier \cite{glorot2010understanding} to initialize the added layers. AdamW~\cite{adam, adamw}  is adopted for training with an initial learning rate of $1\times10^{-4}$ and batch size of 16. Following \cite{retinanet, mask_rcnn}, we employ 1$\times$ training schedule, \ie, training the detection models for 12 epochs. The training images are resized into shorter side of 800 pixels and longer side of no more than 1,333 pixels. For testing, the shorter side of the images is also resized to 800 pixels. The implementation is based on the \texttt{mmdetection} \cite{mmdetection} codebase and the experiments are run on 8 NVIDIA A100 GPUs. 

\myPara{Results}
Equipped with RetinaNet for object detection, PoolFormer-based models consistently outperform their comparable ResNet counterparts as shown in Table \ref{tab:coco_det}. For instance, PoolFormer-S12 achieves 36.2 AP, largely surpassing that of ResNet-18 (31.8 AP). 
Similar results are observed for those models based on Mask R-CNN on object detection and instance segmentation. For example, PoolFormer-S12 largely surpasses ResNet-18 (bounding box AP 37.3 \vs 34.0, and mask AP 34.6 \vs 31.2). 
Overall, for COCO object detection and instance segmentation, PoolForemrs achieve competitive performance, consistently outperforming those counterparts of ResNet.

\begin{table*}[t]
\small
\centering
\setlength{\tabcolsep}{2pt}
\begin{tabular}{l|l|c c c}
\toprule
	Ablation & Variant & Params (M) & MACs (G) & Top-1 (\%) \\
\whline
Baseline & None (PoolFormer-S12) & 11.9 & 1.8 & 77.2 \\
\hline
\multirow{6}{*}{Token mixers}  & Pooling $\rightarrow$ Identity mapping & 11.9 & 1.8 & 74.3 \\
             & Pooling $\rightarrow$ Global random matrix$^*$ (extra 21M frozen
parameters) & 11.9 & 3.3 & 75.8 \\
             & Pooling $\rightarrow$ Depthwise Convolution \cite{chollet2017xception, mamalet2012simplifying} & 11.9 & 1.8 & 78.1 \\
             & Pooling size 3 $\rightarrow$ 5 & 11.9 & 1.8 & 77.2 \\
             & Pooling size 3 $\rightarrow$ 7 & 11.9 & 1.8 & 77.1 \\
             & Pooling size 3 $\rightarrow$ 9 & 11.9 & 1.8 & 76.8 \\
\hline
\multirow{3}{*}{Normalization}  & Modified Layer Normalization$^\dag$ $\rightarrow$  Layer Normalization \cite{layer_norm} & 11.9 & 1.8 & 76.5 \\
             & Modified Layer Normalization$^\dag$ $\rightarrow$ Batch Normalization \cite{batch_norm} & 11.9 & 1.8 & 76.4 \\
             & Modified Layer Normalization$^\dag$ $\rightarrow$ None & 11.9 & 1.8 & 46.1 \\
\hline
\multirow{2}{*}{Activation}  & GELU \cite{gelu} $\rightarrow$  ReLU \cite{relu}  & 11.9 & 1.8 & 76.4 \\
             & GELU $\rightarrow$ SiLU \cite{silu} & 11.9 & 1.8 & 77.2 \\
\hline
\multirow{2}{*}{Other components}  & Residual connection \cite{gelu} $\rightarrow$  None  & 11.9 & 1.8 & 0.1 \\
             & Channel MLP $\rightarrow$ None & 2.5 & 0.2 & 5.7 \\
\hline
\multirow{4}{*}{Hybrid Stages}  & [Pool, Pool, Pool, Pool] $\rightarrow$  [Pool, Pool,                                 Pool, Attention] & 14.0 & 1.9 & 78.3 \\
                                & [Pool, Pool, Pool, Pool] $\rightarrow$  [Pool, Pool, Attention, Attention]  & 16.5 & 2.5 & 81.0 \\
                                & [Pool, Pool, Pool, Pool] $\rightarrow$  [Pool, Pool, Pool, SpatialFC] & 11.9 & 1.8 & 77.5  \\
                                & [Pool, Pool, Pool, Pool] $\rightarrow$  [Pool, Pool, SpatialFC, SpatialFC] & 12.2 & 1.9 & 77.9 \\
             
\bottomrule
\end{tabular}
\vspace{-3mm}
\caption{\textbf{Ablation for PoolFormer on ImageNet-1K classification benchmark.} PoolFormer-S12 is utilized as the baseline to conduct ablation study. The top-1 accuracy on the validation set is reported. $^*$This token mixer utilizes global random matrix $W_R\in \mathbb{R}^{N\times N}$ (parameters are frozen after random initialization) to conduct token mixing by $X'=W_RX$ where $X\in \mathbb{R}^{N \times C}$ are input tokens with the token length of $N$ and channel dimension of $C$. $^\dag$Modified Layer Normalization (MLN) computes the mean and variance along token and channel dimensions compared with vanilla Layer Normalization only along channel dimension. MLN can be implemented with GroupNorm API in PyTorch by specifying the group number equal to 1. The numbers of MACs are counted by \texttt{fvcore} \cite{fvcore} library.
}
\label{tab:ablation}
\normalsize
\vspace{-2mm}
\end{table*}

\vspace{-1pt}
\subsection{Semantic segmentation}
\vspace{-3pt}
\myPara{Setup} 
ADE20K~\cite{ade20k}, a challenging scene parsing benchmark, is selected to evaluate the models for semantic segmentation. The dataset includes 20K and 2K images in the training and validation set, respectively, covering 150 fine-grained semantic categories. PoolFormers are evaluated as backbones equipped with Semantic FPN~\cite{fpn}. ImageNet-1K trained checkpoints are used to initialize the backbones while Xavier~\cite{glorot2010understanding} is utilized to initialize other newly added layers. Common practices~\cite{fpn,chen2017deeplab} train models for 80K iterations with a batch size of 16. To speed up training, we double the batch size to 32 and decrease the iteration number to 40K. The AdamW \cite{adam, adamw} is employed with an initial learning rate of $2\times 10^{-4}$ that will decay in the polynomial decay schedule with a power of 0.9. Images are resized and cropped into $512\times 512$ for training and are resized to shorter side of 512 pixels for testing. Our implementation is based on the \texttt{mmsegmentation} \cite{mmseg2020} codebase and the experiments are conducted on 8 NVIDIA A100 GPUs.

\myPara{Results} Table~\ref{tab:ade20k} shows the ADE20K semantic segmentation performance of different backbones using FPN~\cite{fpn}. PoolFormer-based models consistently outperform the models with backbones of CNN-based ResNet~\cite{resnet} and ResNeXt \cite{xie2017aggregated} as well as Transformer-based PVT. For instance, PoolFormer-12 achieves mIoU of 37.1, 4.3 and 1.5 better than ResNet-18 and PVT-Tiny, respectively.

These results demonstrate that our PoorFormer which serves as backbone can attain competitive performance on semantic segmentation although it only utilizes pooling for basically communicating information among tokens. This further indicates the great potential of MetaFormer and supports our claim that MetaFormer is actually what we need.

\subsection{Ablation studies}
\label{sec:ablation}
The experiments of ablation studies are conducted on ImageNet-1K \cite{imagenet}.
Table \ref{tab:ablation} reports the ablation study of PoolFormer. 
We discuss the ablation below according to the following aspects.

\myPara{Token mixers} Compared with Transformers, the main change made by PoolFormer is using simple pooling as a token mixer. We first conduct ablation for this operator by directly replacing pooling with identity mapping. Surprisingly, MetaFormer with identity mapping can still achieve 74.3\% top-1 accuracy, supporting the claim that MetaFormer is actually what we need to guarantee reasonable performance.

Then the pooling is replaced with global random matrix $W_R\in \mathbb{R^{N\times N}}$ for each block. The matrix is initialized with random values from a uniform distribution on the interval [0, 1), and then Softmax is utilized to normalize each row. After random initialization, the matrix parameters are frozen and it conducts token mixing by $X'=W_RX$ where $X\in \mathbb{R^{N\times C}}$ are the input token features with the token length of $N$ and channel dimension of $C$. The token mixer of random matrix introduces extra 21M frozen parameters for the S12 model since the token lengths are extremely large at the first stage. Even with such random token mixing method, the model can still achieve reasonable performance of 75.8\% accuracy, 1.5\% higher than that of identity mapping. It shows that MetaFormer can still work well even with random token mixing, not to say with other well-designed token mixers.

Further, pooling is replaced with Depthwise Convolution \cite{chollet2017xception, mamalet2012simplifying} that has learnable parameters for spatial modeling. Not surprisingly, the derived model still achieve highly competitive performance with top-1 accuracy of 78.1\%, 0.9\% higher than PoolFormer-S12 due to its better local spatial modeling ability. Until now, we have specified multiple token mixers in Metaformer, and all resulted models keep promising results, well supporting the claim that MetaFormer is the key to guaranteeing models' competitiveness. Due to the simplicity of pooling, it is mainly utilized as a tool to demonstrate MetaFormer.

We test the effects of pooling size on PoolFormer. We observe similar performance when pooling sizes are 3,  5, and 7. However, when the pooling size increases to 9, there is an obvious performance drop of 0.5\%.  Thus, we adopt the default pooing size of 3 for PoolFormer.

\myPara{Normalization} We modify Layer Normalization \cite{layer_norm} into Modified Layer Normalization (MLN) that computes the mean and variance along token and channel dimensions compared with only channel dimension in vanilla Layer Normalization. The shape of learnable affine parameters of MLN keeps the same as that of Layer Normalization, \ie, $\mathbb{R^C}$. MLN can be implemented with GroupNorm API in PyTorch by setting the group number as 1. See the appendix for details. We find PoolFormer prefers MLN with 0.7\% or 0.8\% higher than Layer Normalization or Batch Normalization. Thus, MLN is set as default for PoolFormer. When removing normalization, the model can not be trained to converge well, and its performance dramatically drops to only 46.1\%.

\myPara{Activation} We change GELU \cite{gelu} to ReLU \cite{relu} or SiLU \cite{silu}. When ReLU is adopted for activation, an obvious performance drop of 0.8\% is observed. For SiLU, its performance is almost the same as that of GELU. Thus, we still adopt GELU as default activation.

\myPara{Other components} Besides token mixer and normalization discussed above, residual connection \cite{resnet} and channel MLP \cite{rosenblatt1961principles, rumelhart1985learning} are two other important components in MetaFormer. Without residual connection or channel MLP, the model cannot converge and only achieves the accuracy of 0.1\%/5.7\%, proving the indispensability of these parts.

\myPara{Hybrid stages} Among token mixers based on pooling, attention, and spatial MLP, the pooling-based one can handle much longer input sequences while attention and spatial MLP are good at capturing global information. 
Therefore, it is intuitive to stack MetaFormers with pooling in the bottom stages to handle long sequences and use attention or spatial MLP-based mixer in the top stages, considering the sequences have been largely shortened. Thus, we replace the token mixer pooling with attention or spatial FC \footnote{Following \cite{resmlp}, we use only one spatial fully connected layer as a token mixer, so we call it FC.} in the top one or two stages in PoolFormer. From Table \ref{tab:ablation}, the hybrid models perform quite well. The variant with pooling in the bottom two stages and attention in the top two stages delivers highly competitive performance. 
It achieves 81.0\% accuracy with only 16.5M parameters and 2.5G MACs. 
As a comparison, ResMLP-B24 needs $7.0\times$ parameters (116M) and $9.2\times$ MACs (23.0G) to achieve the same accuracy. These results indicate that combining pooling with other token mixers for MetaFormer may be a promising direction to further improve the performance.

\section{Conclusion and future work}
In this work, we abstracted the attention in Transformers as a token mixer, and the overall Transformer as a general architecture termed MetaFormer where the token mixer is not specified. Instead of focusing on specific token mixers, we point out that MetaFormer is actually what we need to guarantee achieving reasonable performance. To verify this, we deliberately specify token mixer as extremely simple pooling for MetaFormer. It is found that the derived PoolFormer model can achieve competitive performance on different vision tasks, which well supports that ``MetaFormer is actually what you need for vision". 

In the future, we will further evaluate PoolFormer under more different learning settings, such as self-supervised learning and transfer learning. Moreover, it is interesting to see whether PoolFormer still works on NLP tasks to further support the claim ``MetaFormer is actually what you need" in the NLP domain. We hope that this work can inspire more future research devoted to improving the fundamental architecture MetaFormer instead of paying too much attention to the token mixer modules. 

\section*{Acknowledgement}
The authors would like to thank Quanhong Fu at Sea AI Lab for the help to improve the technical writing aspect of this paper. Weihao Yu would like to thank TPU Research Cloud (TRC) program and Google Cloud research credits for the support of partial computational resources.
This project is in part supported by NUS Faculty Research Committee Grant~(WBS: A-0009440-00-00). Shuicheng Yan and Xinchao Wang are the corresponding authors.

{\small
\bibliographystyle{ieee_fullname}
\bibliography{6_references}
}

\newpage
\appendix
\section{Detailed hyper-parameters on ImageNet-1K}

\myPara{PoolFormer}
On ImageNet-1K classification benchmark, we utilize the hyper-parameters shown in Table \ref{tab:hyperparameter} to train models in our paper. Based on the relation between batch size and learning rate in Table \ref{tab:hyperparameter}, we set the batch size as 4096 and learning rate as $4\times 10^{-3}$. For stochastic depth, following the original paper \cite{stochastic_depth}, we linearly increase the probability of dropping a layer from 0.0 for the bottom block to $d_r$ for the top block.

\begin{table*}[htbp]
\centering
\begin{tabular}{@{}l|ccccc@{}}
\toprule
 & \multicolumn{5}{c}{PoolFormer} \\ 
 & S12 & S24 & S36 & M36 & M48 \\
\midrule
Peak drop rate of stoch. depth $d_r$ & 0.1 & 0.1 & 0.2 & 0.3 & 0.4 \\
LayerScale initialization $\epsilon$ & $10^{-5}$ & $10^{-5}$ & $10^{-6}$ & $10^{-6}$ & $10^{-6}$ \\
\hline
Data augmentation & \multicolumn{5}{c}{AutoAugment} \\
Repeated Augmentation & \multicolumn{5}{c}{off} \\
Input resolution & \multicolumn{5}{c}{224} \\
Epochs & \multicolumn{5}{c}{300} \\
Warmup epochs & \multicolumn{5}{c}{5} \\
Hidden dropout & \multicolumn{5}{c}{0} \\
GeLU dropout & \multicolumn{5}{c}{0} \\
Classification dropout & \multicolumn{5}{c}{0} \\
Random erasing prob & \multicolumn{5}{c}{0.25} \\
EMA decay & \multicolumn{5}{c}{0} \\
Cutmix $\alpha$ & \multicolumn{5}{c}{1.0} \\
Mixup $\alpha$ & \multicolumn{5}{c}{0.8} \\
Cutmix-Mixup switch prob & \multicolumn{5}{c}{0.5} \\
Label smoothing & \multicolumn{5}{c}{0.1} \\
\tabincell{l}{Relation between peak learning \\ \qquad rate and batch size} & \multicolumn{5}{c}{$\mathrm{lr} = \frac{\mathrm{batch\_size}}{1024}\times 10^{-3}$} \\
Batch size used in the paper & \multicolumn{5}{c}{4096} \\
Peak learning rate used in the paper & \multicolumn{5}{c}{$4 \times 10^{-4}$} \\
Learning rate decay & \multicolumn{5}{c}{cosine} \\
Optimizer & \multicolumn{5}{c}{AdamW} \\
Adam $\epsilon$ & \multicolumn{5}{c}{1e-8} \\
Adam $(\beta_1, \beta_2)$ & \multicolumn{5}{c}{(0.9, 0.999)} \\
Weight decay & \multicolumn{5}{c}{0.05} \\
Gradient clipping & \multicolumn{5}{c}{None} \\
\bottomrule
\end{tabular}
\vspace{-2mm}
\caption{\textbf{Hyper-parameters for image classification on ImageNet-1K}
\label{tab:hyperparameter}
}
\end{table*}

\myPara{Hybrid Models}
We use the hyper-parameters for all models except for 
the hybrid models with token mixers of pooling and attention. 
For these hybrid models, we find that
they achieve much better performances
by setting batch size as 1024, 
learning rate as $10^{-3}$,
and normalization as
Layer Normalization \cite{layer_norm}.

\section{Training for longer epochs}
In our paper, PoolFormer models are trained for the default 300 epochs on ImageNet-1K. For DeiT \cite{deit}/ResMLP\cite{resmlp}, it is observed that the performance saturates after 400/800 epochs. Thus, we also conduct the experiments of training longer for PoolFormer-S12 and the results are shown in Table \ref{tab:long_epochs}. We observe that PoolFormer-S12 obtains saturated performance after around 2000 epochs with a top-1 accuracy improvement of 1.8\%. However, for fair comparison with other ViT/MLP-like models, we still train PoolFormers for 300 epochs by default.

\begin{table*}[t]
\centering
\setlength{\tabcolsep}{10pt}
\begin{tabular}{l c c c c c c c c}
\toprule
\# Epochs & 300 (default) & 400 & 500 & 1000 & 1500 & 2000 & 2500 & 3000 \\
\midrule
PoolFormer-S12 & 77.2 & 77.5 & 77.9 & 78.4 & 78.6 & 78.8 & 78.8 & 78.8 \\
\bottomrule
\end{tabular}
\vspace{-2mm}
\caption{\textbf{Performance of PoolFormer trained for different numbers of epochs}.}
\label{tab:long_epochs} 
\end{table*}

\section{Qualitative results}
We use Grad-CAM \cite{gradcam} to visualize the results of different models trained on ImageNet-1K. We find that although ResMLP \cite{resmlp} also activates some irrelevant parts, all models can locate the semantic objects. The activation parts of DeiT \cite{deit} and ResMLP \cite{resmlp} in the maps are more scattered, while those of RSB-ResNet \cite{resnet_improved, resnet} and PoolFormer are more gathered.

\begin{figure*}[t]
    \centering
    \begin{subfigure}[b]{0.19\textwidth}
        \centering
        \includegraphics[width=1\textwidth]{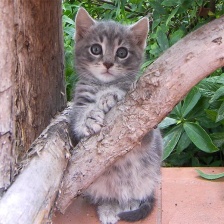}
        \includegraphics[width=1\textwidth]{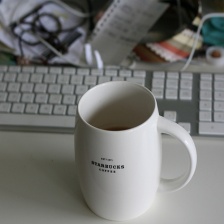}
        \includegraphics[width=1\textwidth]{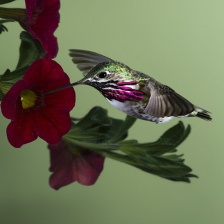}
        \includegraphics[width=1\textwidth]{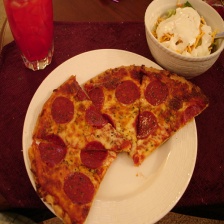}
    \end{subfigure}    
    \begin{subfigure}[b]{0.19\textwidth}
        \centering
        \includegraphics[width=1\textwidth]{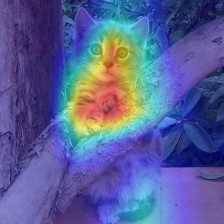}
        \includegraphics[width=1\textwidth]{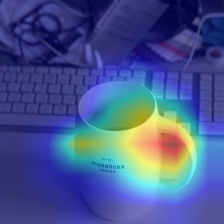}
        \includegraphics[width=1\textwidth]{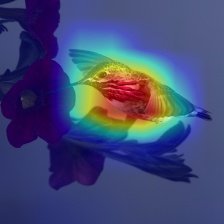}
        \includegraphics[width=1\textwidth]{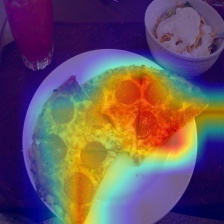}
    \end{subfigure}  
    \begin{subfigure}[b]{0.19\textwidth}
        \centering
        \includegraphics[width=1\textwidth]{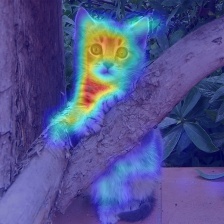}
        \includegraphics[width=1\textwidth]{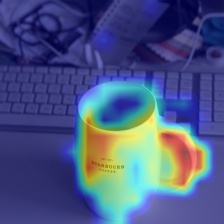}
        \includegraphics[width=1\textwidth]{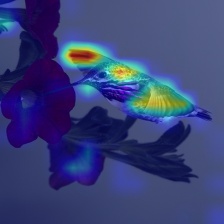}
        \includegraphics[width=1\textwidth]{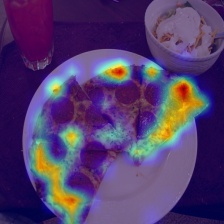}
    \end{subfigure}  
    \begin{subfigure}[b]{0.19\textwidth}
        \centering
        \includegraphics[width=1\textwidth]{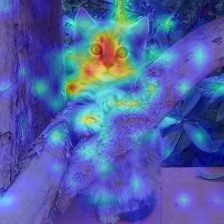}
        \includegraphics[width=1\textwidth]{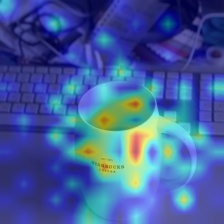}
        \includegraphics[width=1\textwidth]{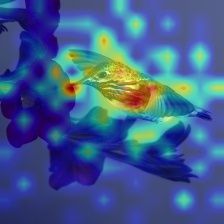}
        \includegraphics[width=1\textwidth]{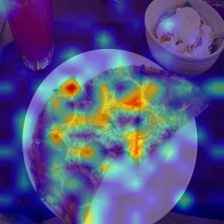}
    \end{subfigure}  
    \begin{subfigure}[b]{0.19\textwidth}
        \centering
        \includegraphics[width=1\textwidth]{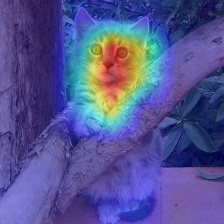}
        \includegraphics[width=1\textwidth]{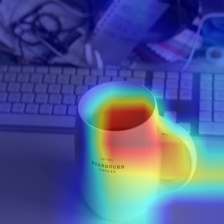}
        \includegraphics[width=1\textwidth]{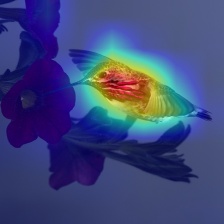}
        \includegraphics[width=1\textwidth]{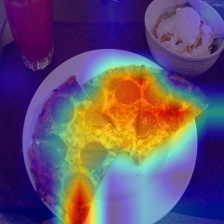}
    \end{subfigure}  
    \begin{center}
    	 ~~~~~~~Input\qquad \qquad \quad RSB-ResNet-50 \cite{resnet_improved} \qquad ~~ DeiT-small \cite{deit} \qquad ~~ ResMLP-S24 \cite{resmlp} \qquad ~~ PoolFormer-S24
    \end{center}  
    \caption{
        \label{grad_cam} Grad-CAM \cite{gradcam} activation maps of the models trained on ImageNet-1K. The visualized images are from validation set. 
    }
\end{figure*}
\section{Comparison between Layer Normalization and Modified Layer Normalization}
We modify Layer Normalization \cite{layer_norm} into Modified Layer Normalization (MNN). It computes the mean and variance along spatial and channel dimensions, compared with only channel dimension in vanilla Layer Normalization. The shape of learnable affine parameters of MLN keeps the same as that of Layer Normalization, \ie, $\mathbb{R^C}$. MLN can be implemented with GroupNorm API in PyTorch by setting the group number as 1. The comparison details are shown in Algorithm \ref{alg:norm}.

\begin{algorithm*}[t]
\caption{Comparison between Layer Normalization and Modified Layer Normalization, PyTorch-like Code}
\label{alg:norm}
\definecolor{codeblue}{rgb}{0.25,0.5,0.5}
\definecolor{codekw}{rgb}{0.85, 0.18, 0.50}
\lstset{
  backgroundcolor=\color{white},
  basicstyle=\fontsize{7.5pt}{7.5pt}\ttfamily\selectfont,
  columns=fullflexible,
  breaklines=true,
  captionpos=b,
  commentstyle=\fontsize{7.5pt}{7.5pt}\color{codeblue},
  keywordstyle=\fontsize{7.5pt}{7.5pt}\color{codekw},
}
\begin{lstlisting}[language=python]
import torch.nn as nn


class LayerNormChannel(nn.Module):
    """
    Vanilla Layer Normalization normalizes vectors along channel dimension.
    Input: tensor in shape [B, C, H, W].
    """
    def __init__(self, num_channels, eps=1e-05):
        super().__init__()
        # The shape of  learnable affine parameters is [num_channels, ].
        self.weight = nn.Parameter(torch.ones(num_channels))
        self.bias = nn.Parameter(torch.zeros(num_channels))
        self.eps = eps

    def forward(self, x):
        u = x.mean(1, keepdim=True) # Compute the means along channel dimension.
        s = (x - u).pow(2).mean(1, keepdim=True) # Compute the variances along channel dimension.
        x = (x - u) / torch.sqrt(s + self.eps)
        x = self.weight.unsqueeze(-1).unsqueeze(-1) * x \
            + self.bias.unsqueeze(-1).unsqueeze(-1)
        return x
        

class ModifiedLayerNorm(nn.Module):
    """
    Modified Layer Normalization normalizes vectors along channel dimension and spatial dimensions.
    Input: tensor in shape [B, C, H, W]
    """
    def __init__(self, num_channels, eps=1e-05):
        super().__init__()
        # The shape of  learnable affine parameters is also [num_channels, ], keeping the same as vanilla Layer Normalization.
        self.weight = nn.Parameter(torch.ones(num_channels))
        self.bias = nn.Parameter(torch.zeros(num_channels))
        self.eps = eps

    def forward(self, x):
        u = x.mean([1, 2, 3], keepdim=True) # Compute the mean along channel dimension and spatial dimensions.
        s = (x - u).pow(2).mean([1, 2, 3], keepdim=True) # Compute the variance along channel dimension and spatial dimensions.
        x = (x - u) / torch.sqrt(s + self.eps)
        x = self.weight.unsqueeze(-1).unsqueeze(-1) * x \
            + self.bias.unsqueeze(-1).unsqueeze(-1)
        return x
        

# Modified Layer Normalization can also be implemented using GroupNorm API in PyTorch by setting the group number as 1.        
class ModifiedLayerNorm(nn.GroupNorm):
    """
    Modified Layer Normalization implemented by Group Normalization with 1 group.
    Input: tensor in shape [B, C, H, W]
    """
    def __init__(self, num_channels, **kwargs):
        super().__init__(1, num_channels, **kwargs)
\end{lstlisting}
\end{algorithm*}

\section{Code in PyTorch}
We provide the PyTorch-like code in Algorithm \ref{alg:module} associated with the modules used in the PoolFormer block. Algorithm~\ref{alg:block} further shows the PoolFormer block built with these modules.

\begin{algorithm*}[t]
\caption{Modules for PoolFormer block, PyTorch-like Code}
\label{alg:module}
\definecolor{codeblue}{rgb}{0.25,0.5,0.5}
\definecolor{codekw}{rgb}{0.85, 0.18, 0.50}
\lstset{
  backgroundcolor=\color{white},
  basicstyle=\fontsize{7.5pt}{7.5pt}\ttfamily\selectfont,
  columns=fullflexible,
  breaklines=true,
  captionpos=b,
  commentstyle=\fontsize{7.5pt}{7.5pt}\color{codeblue},
  keywordstyle=\fontsize{7.5pt}{7.5pt}\color{codekw},
}
\begin{lstlisting}[language=python]
import torch.nn as nn


class ModifiedLayerNorm(nn.GroupNorm):
    """
    Modified Layer Normalization implemented by Group Normalization with 1 group.
    Input: tensor in shape [B, C, H, W]
    """
    def __init__(self, num_channels, **kwargs):
        super().__init__(1, num_channels, **kwargs)


class Pooling(nn.Module):
    """
    Implementation of pooling for PoolFormer
    --pool_size: pooling size
    Input: tensor with shape [B, C, H, W]
    """
    def __init__(self, pool_size=3):
        super().__init__()
        self.pool = nn.AvgPool2d(
            pool_size, stride=1, padding=pool_size//2, count_include_pad=False)

    def forward(self, x):
        # Subtraction of the input itself is added 
        # since the block already has a residual connection.
        return self.pool(x) - x


class Mlp(nn.Module):
    """
    Implementation of MLP with 1*1 convolutions.
    Input: tensor with shape [B, C, H, W]
    """
    def __init__(self, in_features, hidden_features=None, 
                 out_features=None, act_layer=nn.GELU, drop=0.):
        super().__init__()
        out_features = out_features or in_features
        hidden_features = hidden_features or in_features
        self.fc1 = nn.Conv2d(in_features, hidden_features, 1)
        self.act = act_layer()
        self.fc2 = nn.Conv2d(hidden_features, out_features, 1)
        self.drop = nn.Dropout(drop)
        self.apply(self._init_weights)

    def _init_weights(self, m):
        if isinstance(m, nn.Conv2d):
            trunc_normal_(m.weight, std=.02)
            if m.bias is not None:
                nn.init.constant_(m.bias, 0)

    def forward(self, x):
        x = self.fc1(x)
        x = self.act(x)
        x = self.drop(x)
        x = self.fc2(x)
        x = self.drop(x)
        return x
\end{lstlisting}
\end{algorithm*}

\begin{algorithm*}[t]
\caption{PoolFormer block, PyTorch-like Code}
\label{alg:block}
\definecolor{codeblue}{rgb}{0.25,0.5,0.5}
\definecolor{codekw}{rgb}{0.85, 0.18, 0.50}
\lstset{
  backgroundcolor=\color{white},
  basicstyle=\fontsize{7.5pt}{7.5pt}\ttfamily\selectfont,
  columns=fullflexible,
  breaklines=true,
  captionpos=b,
  commentstyle=\fontsize{7.5pt}{7.5pt}\color{codeblue},
  keywordstyle=\fontsize{7.5pt}{7.5pt}\color{codekw},
}
\begin{lstlisting}[language=python]
import torch.nn as nn


class PoolFormerBlock(nn.Module):
    """
    Implementation of one PoolFormer block.
    --dim: embedding dim
    --pool_size: pooling size
    --mlp_ratio: mlp expansion ratio
    --act_layer: activation
    --norm_layer: normalization
    --drop: dropout rate
    --drop path: Stochastic Depth, 
        refer to https://arxiv.org/abs/1603.09382
    --use_layer_scale, --layer_scale_init_value: LayerScale, 
        refer to https://arxiv.org/abs/2103.17239
    """
    def __init__(self, dim, pool_size=3, mlp_ratio=4., 
                 act_layer=nn.GELU, norm_layer=ModifiedLayerNorm, 
                 drop=0., drop_path=0., 
                 use_layer_scale=True, layer_scale_init_value=1e-5):

        super().__init__()

        self.norm1 = norm_layer(dim)
        self.token_mixer = Pooling(pool_size=pool_size)
        self.norm2 = norm_layer(dim)
        mlp_hidden_dim = int(dim * mlp_ratio)
        self.mlp = Mlp(in_features=dim, hidden_features=mlp_hidden_dim, 
                       act_layer=act_layer, drop=drop)

        # The following two techniques are useful to train deep PoolFormers.
        self.drop_path = DropPath(drop_path) if drop_path > 0. \
            else nn.Identity()
        self.use_layer_scale = use_layer_scale
        if use_layer_scale:
            self.layer_scale_1 = nn.Parameter(
                layer_scale_init_value * torch.ones(dim), requires_grad=True)
            self.layer_scale_2 = nn.Parameter(
                layer_scale_init_value * torch.ones(dim), requires_grad=True)

    def forward(self, x):
        if self.use_layer_scale:
            x = x + self.drop_path(
                self.layer_scale_1.unsqueeze(-1).unsqueeze(-1)
                * self.token_mixer(self.norm1(x)))
            x = x + self.drop_path(
                self.layer_scale_2.unsqueeze(-1).unsqueeze(-1)
                * self.mlp(self.norm2(x)))
        else:
            x = x + self.drop_path(self.token_mixer(self.norm1(x)))
            x = x + self.drop_path(self.mlp(self.norm2(x)))
        return x
\end{lstlisting}
\end{algorithm*}

\end{document}